\documentclass{article}

\usepackage[final]{corl_2026}

\usepackage{url}
\usepackage[small]{caption}
\usepackage{graphicx}
\usepackage{wrapfig}
\usepackage{amsmath}
\usepackage{amsthm}
\usepackage{booktabs}
\usepackage{algorithm}
\usepackage{algorithmic}
\usepackage{multirow}

\title{One-Step World Models for Multi-Modal Goal Navigation}

\author{
  Wangtian Shen\textsuperscript{1}, Ziyang Meng\textsuperscript{1}\thanks{Corresponding author.},
  Jinming Ma\textsuperscript{2}, Mingliang Zhou\textsuperscript{2} and
  Diyun Xiang\textsuperscript{2}\\[0.3em]
  \normalfont\textsuperscript{1}Tsinghua University,
  \textsuperscript{2}Xiaomi (China)\\[0.3em]
  \normalfont\small
  \texttt{swt22@mails.tsinghua.edu.cn},
  \texttt{ziyangmeng@tsinghua.edu.cn},
  \texttt{majinming3@xiaomi.com}\\
  \texttt{zhoumingliang@xiaomi.com},
  \texttt{xiangdiyun@gmail.com}
}

\allowdisplaybreaks

\begin{document}

\maketitle

\begin{abstract}
World-model-based planning is a promising direction for robot navigation because it enables explicit prediction of future observations under candidate actions. However, existing navigation world models typically rely on multi-step diffusion and autoregressive frame generation, which substantially increases model-side latency and complicates closed-loop deployment. In this work, we present a lightweight one-step world model for navigation built on a 3D U-Net with efficient spatial-temporal attention. The model predicts all future frames non-autoregressively in a single pass, while pretraining and random trajectory sampling improve performance in limited-data settings. We integrate this world model into a unified planning framework that supports image-goal, language-goal, and point-goal navigation through modality-specific scoring functions and anchor-based trajectory initialization. Extensive closed-loop experiments in both simulation and real-world environments show that the proposed approach improves generation efficiency and achieves strong navigation performance against state-of-the-art baselines.
Project page: \url{https://robotnav-bot.github.io/nav-onestepwm/}.

\end{abstract}


\section{Introduction}

Visual navigation constitutes a cornerstone of autonomous robotics, enabling robots to perceive their surroundings and reach target goals in complex, unstructured environments. In recent years, by leveraging large-scale datasets, methods like GNM~\cite{shah2023gnm}, ViNT~\cite{shah2023vint} and NoMaD~\cite{sridhar2024nomad} have demonstrated remarkable robustness and generalization capabilities. However, these end-to-end policies often operate as ``black boxes'', mapping observations directly to actions without explicit 3D spatial reasoning, which limits their performance in challenging navigation tasks.

World models~\cite{ha2018world} offer an appealing alternative by predicting future observations conditioned on actions, thereby supporting planning with richer spatial feedback~\cite{yang2025mindjourney}. Recent works have begun to combine video prediction with optimization-based planning~\cite{bar2025navigation,zhou2025learning}, but their practical use is constrained by computational cost. Existing navigation world models often rely on multi-step diffusion and autoregressive frame-by-frame generation, so planning requires repeatedly invoking expensive generative models over many candidate trajectories. This makes the model-side cost of planning unnecessarily high and limits the practicality of closed-loop execution. In addition, prior navigation world model evaluations mostly focus on image-goal settings, leaving open the question of whether a single world-model-based planner can support multiple goal modalities within one unified framework.

\begin{figure}[t]
    \centering
    \includegraphics[width=1\linewidth]{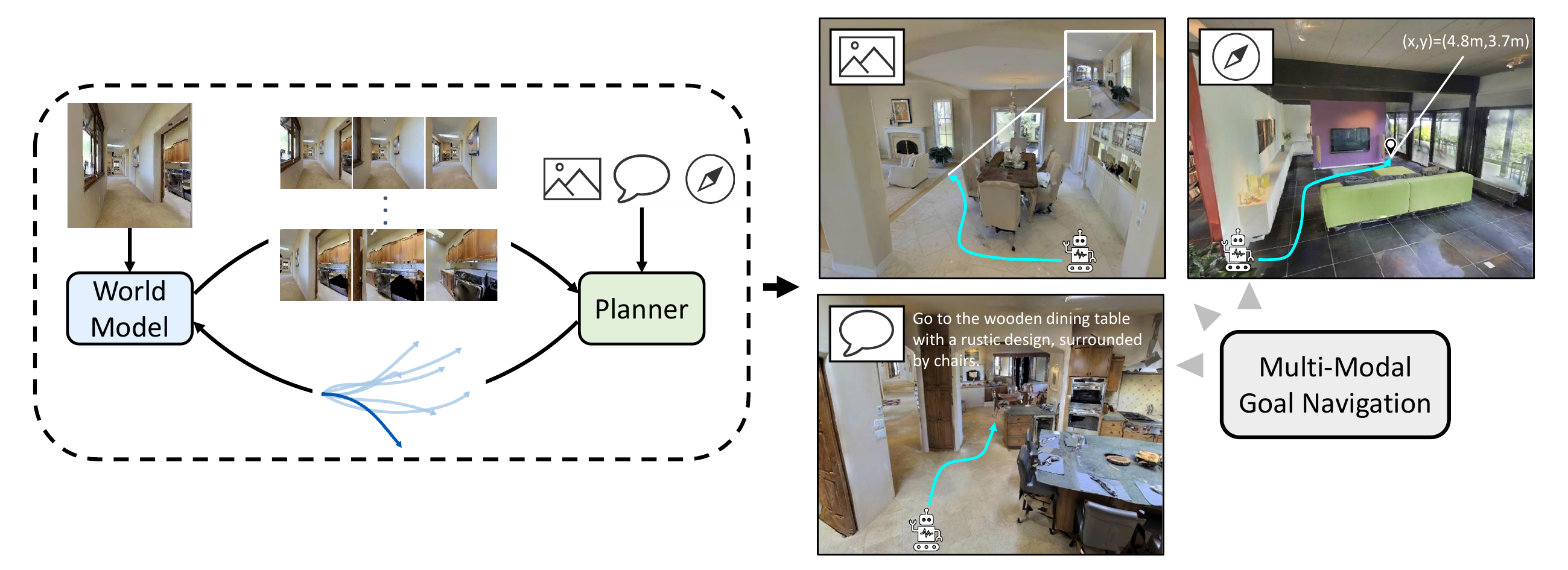}
    \caption{We present a unified navigation framework founded on world models and planning strategies. The proposed system can be directly used to support diverse goal modalities, seamlessly accommodating image-goal, language-goal, and point-goal navigation tasks.}
    \label{fig:intro}
\end{figure}

In this work, we propose a lightweight navigation world model for efficient future prediction. Unlike standard diffusion-based approaches that suffer from slow iterative sampling, our model adopts the one-step generation paradigm~\cite{frans2024one} and a computation-efficient spatial-temporal attention mechanism embedded within a 3D U-Net backbone, enabling non-autoregressive prediction of all future frames. To improve performance in limited-data regimes, we further incorporate pretraining and random trajectory sampling. Building upon this world model, we integrate it with an optimization-based planning framework with task-specific loss functions, enabling unified image-goal, language-goal, and point-goal navigation, as illustrated in Figure~\ref{fig:intro}. Together, these components form a robust closed-loop world-model-based navigation system in simulation and real-world environments.

Our main contributions are summarized as follows:
\begin{itemize}
    \item We propose a lightweight but effective navigation world model leveraging one-step generation paradigm, spatial-temporal attention, and pretraining recipe.
    
    \item We develop a unified navigation system for multi-modal goal navigation by integrating our world model with a planning framework utilizing anchor-based initialization.
    
    \item We demonstrate the superiority of the proposed method through closed-loop experiments in both simulation and the real world.
\end{itemize}

\section{RELATED WORK}
\subsection{Learning Based Visual Navigation}
Traditional solutions to navigation problem typically consist of distinct localization, mapping, and planning modules~\cite{lavalle2006planning}. The advent of machine learning shifts the focus towards learning-based solutions for visual navigation. GNM~\cite{shah2023gnm} shows that training on diverse robotic datasets boosts generalization. ViNT~\cite{shah2023vint} and EffoNAV~\cite{shen2025effonav} further advance this by employing transformer architectures and visual foundation models to establish foundational models for image-goal navigation in general environments. NoMaD~\cite{sridhar2024nomad} leverages diffusion model to execute both navigation and exploration tasks within a single framework. NavDP~\cite{cai2025navdp} enables robust and efficient sim-to-real navigation by leveraging diffusion-based trajectory generation alongside a critic function. LeLaN~\cite{hirose2024lelan} learns language-conditioned navigation policies from unlabeled and action-free videos. OmniVLA~\cite{hirose2025omnivla} presents a generalizable and flexible omni-modal robotic foundation model. Although significant progress has been made by introducing learning-based solutions for navigation, current methods exhibit limited spatial reasoning abilities, leading to suboptimal navigation performance.

\subsection{World Model for Embodied Intelligence}
Unlike end-to-end policy, the primary objective of a world model~\cite{ha2018world} is to simulate environmental dynamics. While early research focused on enhancing sample efficiency in reinforcement learning~\cite{hafner2019dream,hafner2023mastering}, recent works leverage world models for high-dimensional visual generation and planning. Approaches like Pathdreamer~\cite{koh2021pathdreamer}, UniPi~\cite{du2023learning}, and Genie~\cite{bruce2024genie} generate plausible future observations to facilitate interaction and navigation. Furthermore, integrating world models with planning has shown significant promise. NWM~\cite{bar2025navigation} utilizes generative predictions to optimize action sequences, while~\cite{zhou2025learning} introduces 3D memory for temporal consistency. However, the high computational cost of these generative processes makes real-time deployment impossible. On the other hand, approaches like DINO-WM~\cite{zhou2024dino} and~\cite{zhang2025latent} construct world models within the latent space to minimize computational overhead. However, performing generation and planning solely in the latent space often cannot produce explicit 3D spatial prediction and different goal modals cannot be integrated into a unified framework.

\section{METHODS}

\begin{figure}[t]
    \centering
    \includegraphics[width=0.9\linewidth]{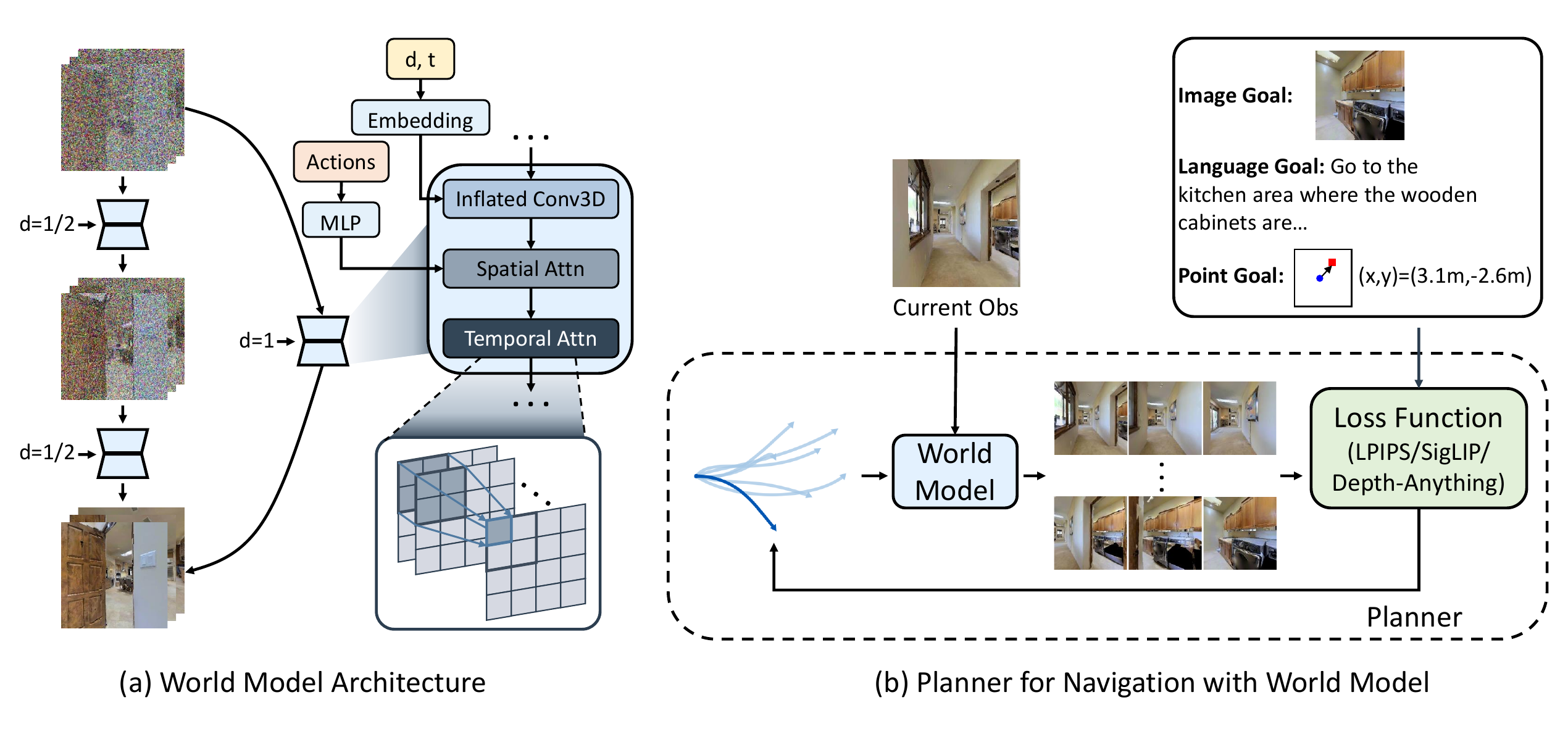}
    \caption{Illustration of our world model architecture and the world model-based planner. (a) Architecture of the proposed navigation world model. Built upon a 3D U-Net backbone, the model integrates convolutional layers with spatial-temporal attention blocks and is trained using a shortcut objective. (b) The action optimization process conditioned on navigation goals. Leveraging the world model, we utilize a loss function to score action candidates, thereby enabling navigation with multi-modal goals.}
    \label{fig:architecture}
\end{figure}

\subsection{One-Step World Model for Navigation}
Formally, the navigation world model predicts the future observation $s_{\tau+1}$ from the current observation $s_{\tau}$ and action $a_{\tau}$, where $\tau$ denotes the current time step. Existing generative methods typically rely on diffusion~\cite{ho2020denoising,song2020denoising} or flow-matching models~\cite{lipman2022flow}, whose iterative sampling incurs high computational cost. To enable one-step generation, we adopt the shortcut training objective~\cite{frans2024one}. Specifically, the world model $f_{\theta}$ predicts the velocity vector $v_t = f_{\theta}(s_{\tau+1}^{(t)}, s_{\tau}, a_{\tau}, t, d)$, where $s_{\tau+1}^{(t)}$ is the noisy future observation and the model is conditioned on both the flow timestep $t$ and step size $d$. For the minimum step size $d_{\min}$, we use the standard flow-matching loss. For larger step sizes $d_{\min}<d<1$, we use a bootstrap consistency loss that distills two smaller substeps of size $d/2$ into a single step, as illustrated in Figure~\ref{fig:architecture}(a). The training objective is given by
\begin{align}
    s_{\tau+1}^{(0)} &\sim \mathcal{N}(0, I), \quad s_{\tau+1}^{(1)} \sim \mathcal{D} (Dataset), \\
    \mathcal{L}(\theta) &= \left\| f_{\theta}(s_{\tau+1}^{(t)}, s_{\tau}, a_{\tau}, t, d) - v_{\text{target}} \right\|^2, \\
    v_{\text{target}} &= \begin{cases}
        s_{\tau+1}^{(1)} - s_{\tau+1}^{(0)}, & \text{if } d = d_{\min}, \\
        \frac{1}{2} (v' + v''), & \text{otherwise,}
    \end{cases} \\
    \text{where } \quad v' &= f_{\theta}(s_{\tau+1}^{(t)}, s_{\tau}, a_{\tau}, t, d/2), \\
    v'' &= f_{\theta}(s_{\tau+1}', s_{\tau}, a_{\tau}, t+\frac{d}{2}, d/2),\\
    s_{\tau+1}' &= s_{\tau+1}^{(t)} + v' \cdot \frac{d}{2}.
\end{align}
By employing this distillation training, the navigation world model learns to generate future observations from gaussian noise within one step.

\subsection{Architecture of Navigation World Model}
We now describe the implementation of the world model $f_{\theta}$. Existing navigation world models~\cite{bar2025navigation} mainly rely on autoregressive, frame-by-frame prediction, which suffers from error accumulation over long horizons and incurs high latency. In contrast, our model uses a 3D U-Net backbone to predict 11 future frames in a single pass. Inspired by Stable Diffusion~\cite{rombach2022high}, it combines CNNs and transformers to model spatiotemporal dependencies, as shown in Figure~\ref{fig:architecture}(a). We first use a pretrained VAE~\cite{rombach2022high} to encode observations into a latent space, generate future observations in that space, and then decode the predicted latents back to pixel images.

\textbf{Action Embedding.} The action input $a_{\tau}$ consists of $N$ relative waypoints $(x, y, \phi)$. To ensure continuity, we encode $\phi$ as $(\cos\phi, \sin\phi)$, forming an $(N, 4)$ vector. This input is projected by a 4-layer MLP into a high-dimensional embedding, which conditions the spatial attention blocks via cross-attention.

\textbf{Spatial and Temporal Attention.} Each block of the 3D U-Net consists of a CNN module, a Spatial Attention module, and a Temporal Attention module. Since transformer complexity scales quadratically with context length, global spatiotemporal attention is computationally prohibitive. Following prior work~\cite{blattmann2023stable,wang2023modelscope}, we decouple spatial and temporal modeling. The CNN and Spatial Attention modules operate on each frame independently to capture intra-frame spatial dependencies, while the Temporal Attention module handles inter-frame information propagation. For temporal attention, we adopt a window-based mechanism~\cite{liu2021swin,gu2023seer} that partitions each frame into $m \times m$ regions and restricts attention to tokens within the same window, reducing context length while preserving spatiotemporal consistency.

\subsection{Training Details}
Following the setup described in~\cite{bar2025navigation}, we collect 500 trajectories within a single indoor environment for training. During model training, we adopt a random trajectory sampling strategy: the position of the final frame is selected randomly, and 10 frames are randomly sampled between the current and final frames to generate a single training instance. This approach effectively increases trajectory diversity within the limited dataset.

In our experiments, we observed that relying solely on data collected in the new scene degraded the prediction quality of the navigation world model. To address this, we first train a base model on a large-scale public navigation dataset~\cite{shah2023gnm} to obtain a pretrained navigation world model. We then fine-tune this pretrained model using the specific data collected from the new environment. Our results demonstrate that this pretraining strategy significantly enhances navigation performance.

\subsection{Navigation with World Model}
\label{sec:navigation_w_wm}

\begin{wrapfigure}[10]{r}{0.46\textwidth}
    \centering
    \includegraphics[width=\linewidth]{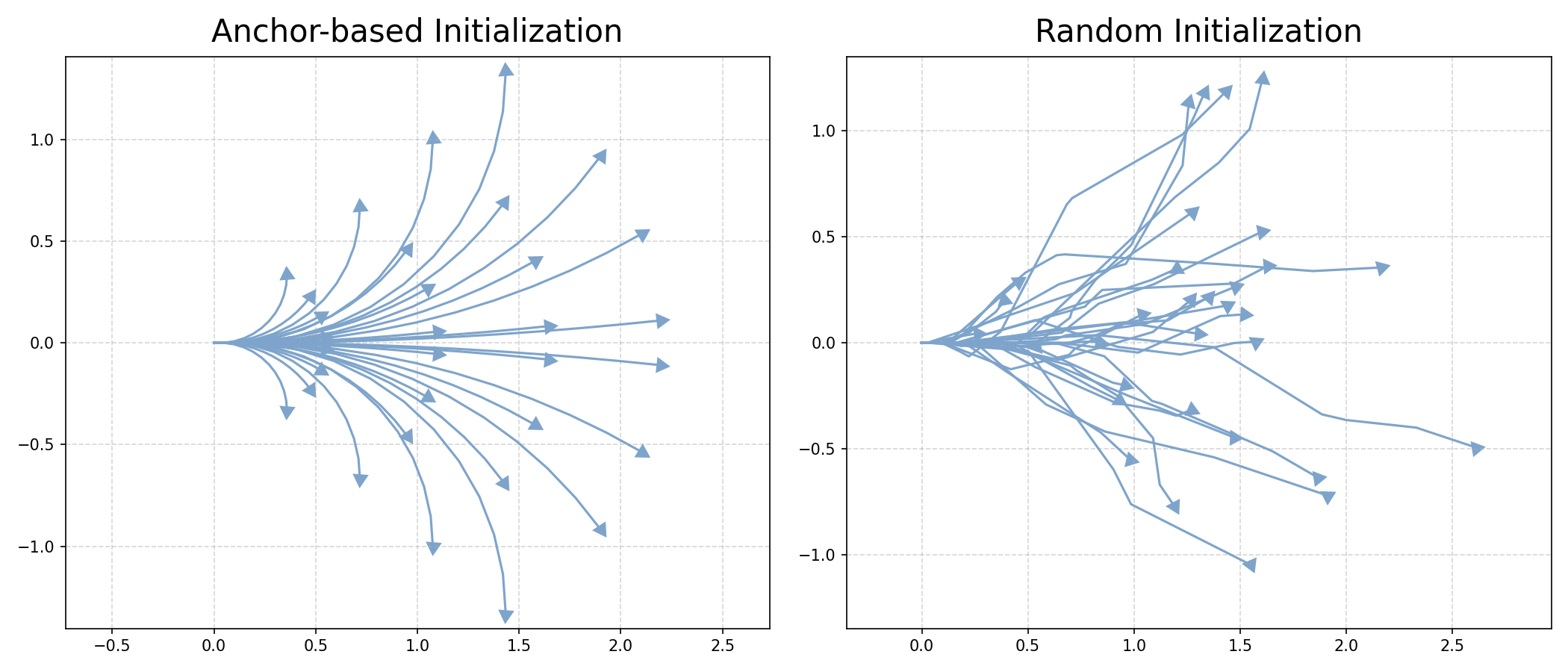}
    \caption{Comparison of anchor-based initialization and random initialization.}
    \label{fig:traj_init}
\end{wrapfigure}

Leveraging the world model's ability to predict future observations conditioned on candidate action trajectories, we implement a model-based planning framework. Specifically, we use the Cross-Entropy Method (CEM)~\cite{rubinstein1997optimization} to optimize candidate trajectories in a sampling--prediction--scoring--selection--update loop, as shown in Figure~\ref{fig:architecture}(b). At each iteration, CEM samples $N$ candidate trajectories, uses the world model to predict their corresponding future observations, scores them with loss functions according to the current navigation goal (e.g., image, language, or point-goal), and retains the top-$K$ elite trajectories to update the sampling distribution. This process is repeated until convergence, after which the best action is executed. A more formal description of the CEM optimization procedure and the modality-specific loss functions is provided in Appendix B.

For all experiments, we limit the sample size to 32 actions at each optimization step. However, we observe that with such a limited sample size, random initialization often fails to sufficiently cover the action space. This results in suboptimal initialization for the planner, ultimately degrading navigation performance. To address this, we employ an anchor-based initialization strategy, where initial candidate trajectories are generated using fixed sets of linear and angular velocities, as shown in Figure~\ref{fig:traj_init}. For each 1-s interval, the anchors use constant linear velocities $v \in \{0.05, 0.1, 0.15, 0.2\}$ m/s and angular velocities $\omega \in \{-0.15, -0.09, -0.05, -0.01, 0.01, 0.05, 0.09, 0.15\}$ rad/s, yielding 32 Cartesian combinations. Empirical results demonstrate that the performance of such an anchor-based approach significantly outperforms that of random initialization.

\section{Generation Experiments}
We evaluate the generation quality of the proposed one-step navigation world model in this section.
\subsection{Experimental Setup}
\label{sec:generation-performance}
We collect training datasets from the MP3D dataset~\cite{chang2017matterport3d}, using the Habitat simulator~\cite{szot2021habitat}. Each dataset consists of 500 trajectories within the respective indoor scene. All models are trained independently on each of the datasets and tested on unseen trajectories.

We compare the proposed method with the following baselines:
\begin{itemize}
    \item \textbf{Original NWM}~\cite{bar2025navigation}. We use the original NWM-B architecture and its diffusion sampler with 250 denoising steps.
    \item \textbf{NWM-Shortcut}. This baseline uses the same NWM-B architecture but replaces its original diffusion sampler with one-step shortcut generation~\cite{frans2024one}, consistent with the generation paradigm used in our method.
    \item \textbf{Ours (w/o random traj)}. An ablative variant of our method without random trajectory sampling strategy in training.
    \item \textbf{Ours (w/o pretrain)}. An ablative variant of our method without first training on large-scale public navigation datasets.
\end{itemize}

For a fair comparison, Original NWM and NWM-Shortcut follow the same training protocol as our model, including the training data, public-data pretraining, scene-specific adaptation, and random trajectory sampling.

\subsection{Generation Performance}

\begin{table*}[htbp]
    \centering
    \small
    \setlength{\tabcolsep}{10pt}
    \begin{tabular}{lccccc}
        \toprule
        Method & LPIPS $\downarrow$ & DreamSim $\downarrow$ & FID $\downarrow$ & Params (M) & Time (s) $\downarrow$ \\
        \midrule
        Original NWM        & 0.319 & 0.300 & \textbf{41.809} & 194 & 31.052 \\
        NWM-Shortcut        & 0.401 & 0.336 & 94.850 & 194 & 0.138 \\
        Ours (w/o random traj) & 0.290 & 0.286 & 89.682 & 175 & \textbf{0.076} \\
        Ours (w/o pretrain) & 0.300 & 0.304 & 98.724 & 175 & \textbf{0.076} \\
        Ours                & \textbf{0.267} & \textbf{0.272} & 83.826 & 175 & \textbf{0.076} \\
        \bottomrule
    \end{tabular}
    \caption{Generation results of different methods}
    \label{tab:generation-results}
\end{table*}

We utilize LPIPS~\cite{zhang2018unreasonable}, DreamSim~\cite{fu2023dreamsim}, and FID~\cite{heusel2017gans} to measure the similarity between the generated output and ground truth. Given the importance of generation speed, we report the inference time for generating 11 future frames with a batch size of 1 on an NVIDIA H200 GPU. The experimental results are presented in Table~\ref{tab:generation-results}. Original NWM achieves the lowest FID but requires 31.052 s for generation. In contrast, our method obtains the best LPIPS and DreamSim while reducing generation latency to 0.076 s. Compared with NWM-Shortcut, it improves all reported quality metrics and reduces latency by approximately 45\% with a similar parameter scale. Ablation studies further show that random trajectory sampling and pretraining on public datasets improve model generalization. Additional qualitative visualization results are provided in Appendix A.
\subsection{Ablation Studies}

We ablate the factorized attention modules by removing within-frame spatial attention and by replacing temporal attention with a causal 3D convolution. As shown in Table~\ref{tab:attention-ablation}, the full model performs best across all three metrics, demonstrating the complementary roles of spatial and temporal modeling.

\begin{table}[H]
    \centering
    \caption{Ablation of spatial and temporal attention modules}
    \label{tab:attention-ablation}
    \small
    \setlength{\tabcolsep}{4pt}
    \begin{tabular}{lccc}
        \toprule
        Model & LPIPS $\downarrow$ & DreamSim $\downarrow$ & FID $\downarrow$ \\
        \midrule
        w/o Spatial Attn. & 0.275 & 0.286 & 93.938 \\
        Causal Temp. Conv. & 0.277 & 0.287 & 93.757 \\
        Full & \textbf{0.267} & \textbf{0.272} & \textbf{83.826} \\
        \bottomrule
    \end{tabular}
\end{table}

\section{Navigation Experiments}
\subsection{Experimental Setup}

Leveraging the proposed world model with task-specific loss functions, our method enables navigation guided by multi-modal goals. We successfully execute image-goal, language-goal and point-goal navigation, achieving strong performance across all three goal modalities. For CEM optimization used in the proposed method, we use a sample size of 32, select the top 5 samples per iteration, and perform 5 iterations. The experimental evaluation consists of 150 episodes across 5 different scenes for image-goal, language-goal and point-goal navigation. We employ \textit{Success Rate (SR)} and \textit{Success-weighted Path Length (SPL)} to evaluate navigation performance.

Apart from the compared methods mentioned in Section~\ref{sec:generation-performance}, we also compare the proposed method with several baselines:
\begin{itemize}
    \item \textbf{NoMaD}~\cite{sridhar2024nomad}. A unified diffusion policy utilizing goal masking to handle both exploration and image-goal navigation.
    \item \textbf{OmniVLA}~\cite{hirose2025omnivla}. An omni-modal robot navigation model built upon the OpenVLA~\cite{kim2024openvla} backbone.
    \item \textbf{NavDP}~\cite{cai2025navdp}. A diffusion-based navigation policy with a unified transformer-based architecture that jointly learns trajectory generation and trajectory evaluation.
    \item \textbf{Ours (random init)}. An ablative variant of our method with random initialization in CEM instead of anchor-based initialization, as described in Section~\ref{sec:navigation_w_wm}.
\end{itemize}

For all world-model baselines, we use the same planning configuration, anchor-based initialization, and task-specific scoring functions. For image-goal navigation, all compared methods use the same topological memory and receive local subgoals generated under the same protocol.

\begin{table*}[htbp]
    \centering
    \footnotesize
    \setlength{\tabcolsep}{8pt}
    \resizebox{\textwidth}{!}{%
    \begin{tabular}{lccccccccc}
        \toprule
        \multirow{2}{*}{Method} & \multicolumn{3}{c}{Image-Goal} & \multicolumn{3}{c}{Language-Goal} & \multicolumn{3}{c}{Point-Goal} \\
        \cmidrule(rr){2-4} \cmidrule(rr){5-7} \cmidrule(rr){8-10}
         & SR & SPL & Lat. (s) & SR & SPL & Lat. (s) & SR & SPL & Lat. (s) \\
        \midrule
        NoMaD & 16.67 & 15.13 & 0.097 & -- & -- & -- & -- & -- & -- \\
        OmniVLA & 36.67 & 34.78 & 0.113 & 62.67 & 56.52 & 0.078 & 40.00 & 36.72 & 0.081 \\
        NavDP & 45.33 & 41.41 & 0.107 & -- & -- & -- & 43.67 & 39.51 & 0.107 \\
        Original NWM & 46.67 & 41.84 & 155.656 & 52.67 & 46.34 & 156.183 & 50.67 & 45.22 & 156.258 \\
        NWM-Shortcut & 43.33 & 38.66 & 1.086 & 51.33 & 45.98 & 1.613 & \textbf{52.67} & \textbf{47.31} & 1.688 \\
        \midrule
        Ours (w/o random traj) & 64.67 & 61.51 & 0.765 & 63.33 & 56.91 & 1.179 & 50.67 & 45.17 & 1.290 \\
        Ours (w/o pretrain) & 64.00 & 61.11 & 0.765 & 65.33 & 58.19 & 1.179 & 48.00 & 43.79 & 1.290 \\
        Ours (random init) & 69.33 & 62.12 & 0.765 & 59.33 & 50.27 & 1.179 & 50.00 & 43.05 & 1.290 \\
        Ours & \textbf{72.67} & \textbf{69.10} & 0.765 & \textbf{69.33} & \textbf{60.90} & 1.179 & 50.00 & 46.24 & 1.290 \\
        \bottomrule
    \end{tabular}%
    }
    \caption{Navigation performance and planning latency on image-goal, language-goal, and point-goal tasks}
    \label{tab:sim-nav-results}
\end{table*}

\textbf{Image-Goal Navigation.}
To implement image-goal navigation, we employ LPIPS~\cite{zhang2018unreasonable} as loss function to score sampled trajectories. Drawing inspiration from prior vision-based approaches~\cite{shah2023gnm,sridhar2024nomad}, we utilize a topological memory system to achieve long-horizon navigation. Specifically, the process initiates with the first observation. At each time step, a distance estimation network identifies the nearest node in the topological graph to estimate the current location. Subsequently, the image of the adjacent node is fed into the policy system as the immediate target. 

\textbf{Language-Goal Navigation.}
To implement language-goal navigation, we employ SigLIP~\cite{zhai2023sigmoid} as the scoring function to evaluate sampled trajectories. We utilize Qwen2-VL~\cite{wang2024qwen2} to generate navigational language instructions conditioned on the target image for the task.

\textbf{Point-Goal Navigation.}
To implement point-goal navigation, we incorporate Depth-Anything~\cite{yang2024depth} to perform monocular depth estimation on the predicted future images. We formulate a composite planning loss function to evaluate candidate trajectories with both goal reaching and safety in mind. Specifically, this loss is defined as the weighted sum of two terms: (1) the Euclidean distance between the terminal point of the sampled trajectory and the target coordinate, and (2) the negative mean depth value in the central region of the predicted image, which is introduced as an explicit safety-oriented term to discourage trajectories that move too close to obstacles and to promote obstacle avoidance during planning.

\subsection{Navigation Performance Results}

\begin{figure}[t!]
    \centering
    \includegraphics[width=0.9\linewidth]{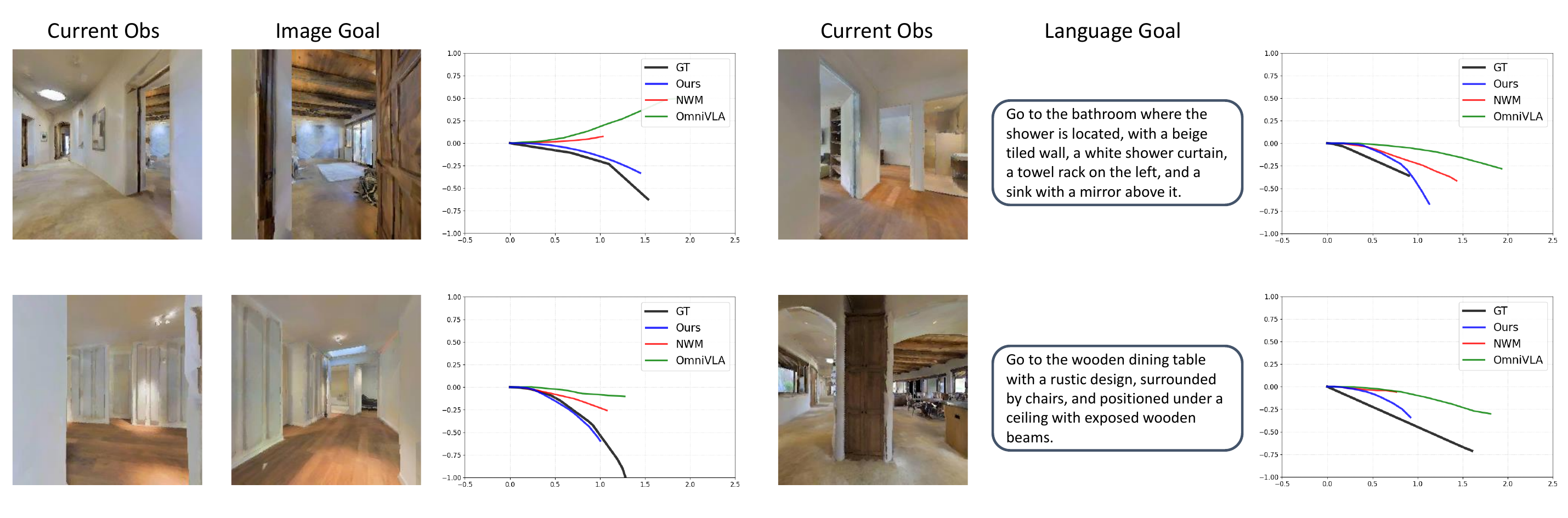}
    \caption{Visualization of trajectories predicted by different methods in image-goal (left) and language-goal (right) navigation.}
    \label{fig:traj}
\end{figure}

The quantitative results in Table~\ref{tab:sim-nav-results} show that the proposed method performs strongly across all three goal modalities. In image-goal navigation, it achieves an SR/SPL of 72.67\%/69.10\%, outperforming Original NWM, NWM-Shortcut, and NavDP. In language-goal navigation, it reaches the highest SR/SPL of 69.33\%/60.90\%, while its point-goal performance remains comparable to both NWM variants and exceeds NavDP. The planning latency of Original NWM exceeds 155 s across all three goal modalities, making it unsuitable for responsive closed-loop replanning.
Figure~\ref{fig:traj} visualizes trajectories generated under identical observations. World-model-based methods exhibit stronger spatial consistency than end-to-end policies when the target is substantially offset from the current viewpoint. In these cases, the proposed approach maintains a more accurate understanding of the goal-relative geometry and produces smoother plans.
\begin{figure}[t]
\centering
\includegraphics[width=0.9\linewidth]{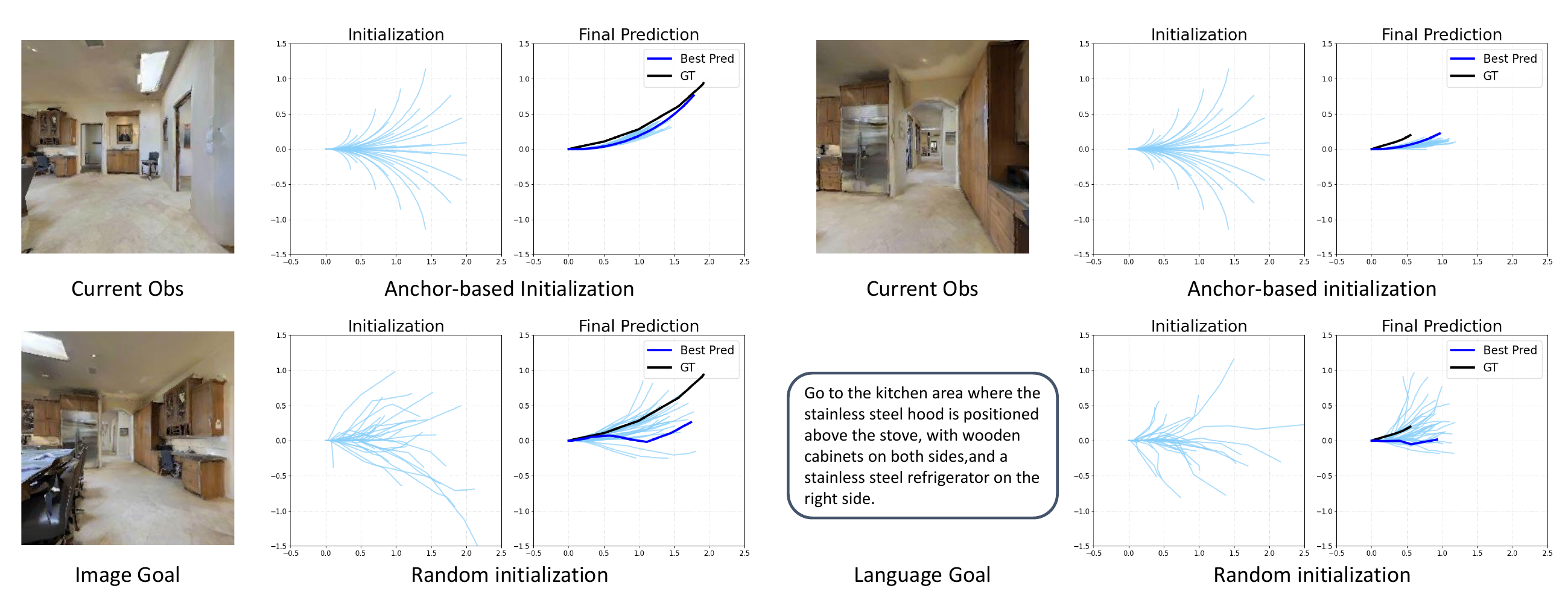}
\caption{Qualitative comparison of predicted trajectories using anchor-based and random initialization, respectively, in image-goal (left) and language-goal (right) navigation.}
\label{fig:random-anchor-plan}
\end{figure}

\newpage
\subsection{Ablations}

\begin{wraptable}[8]{r}{0.42\textwidth}
    \centering
    \small
    \begin{tabular}{lcc}
        \toprule
        Loss Function & SR & SPL\\
        \midrule
        BLIP  & 59.33 & 51.43\\
        CLIP & 61.33 & 53.15\\
        SigLIP (default) & \textbf{69.33} & \textbf{60.90} \\
        \bottomrule
    \end{tabular}
    \caption{Language-goal navigation results of different loss function}
    \label{tab:differnet-loosfunc}
\end{wraptable}

Table~\ref{tab:sim-nav-results} presents the performance comparison between the proposed full method and two ablation variants. The results underscore the critical roles of both the pretraining and random trajectory sampling strategy employed during training. Furthermore, we evaluate the impact of initialization within the CEM planner by comparing random initialization strategy against the proposed anchor-based initialization strategy. As shown in Table~\ref{tab:sim-nav-results}, the anchor-based approach is instrumental in boosting navigation performance. This advantage is further corroborated by the qualitative visualization in Figure~\ref{fig:random-anchor-plan}, which reveals that anchor-based initialization yields smoother and more concentrated trajectory samples, thereby facilitating more effective planning. We further ablate the planning loss used for language-goal navigation in Table~\ref{tab:differnet-loosfunc}. Among the tested vision-language models, SigLIP achieves the best navigation performance, improving both SR and SPL over CLIP and BLIP, and we therefore adopt SigLIP as the default choice.

\section{Real-world Experiments}

\begin{figure}[H]
    \centering
    \includegraphics[width=\linewidth]{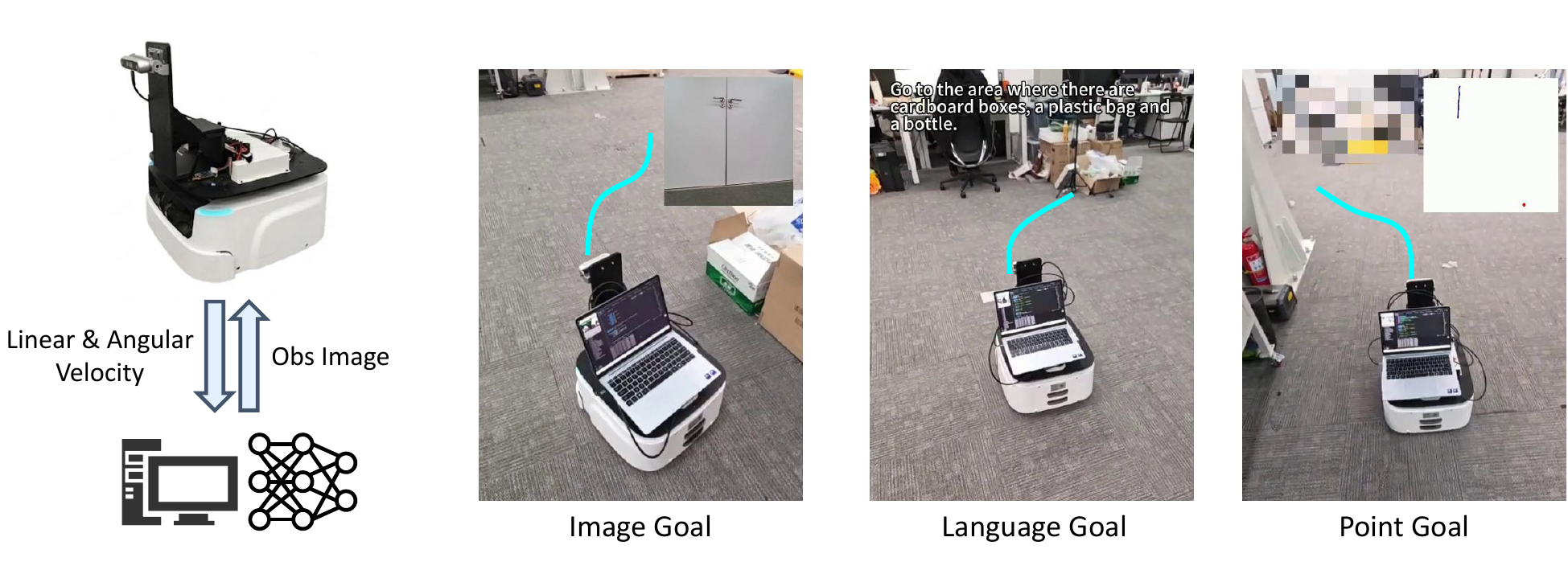}
    \caption{Overview of the robotic platform and visualization of real-world navigation trajectories for image-goal, language-goal, and point-goal tasks.}
    \label{fig:real-world-traj}
\end{figure}

To further validate the effectiveness of the proposed method, we conduct real-world experiments on a mobile robot platform equipped with an Intel RealSense camera, as depicted in Figure~\ref{fig:real-world-traj}. The system operates via a client-server architecture: the robot transmits observations to a cloud server via a network connection, where the model performs inference and returns control commands (i.e., linear and angular velocities). In our real-world deployment, the complete system operates at 1.24 Hz including network communication. We conduct 25 trials for each method on each task. The quantitative results, presented in Table~\ref{tab:real-world-results}, indicate that the proposed approach consistently outperforms baselines in physical environments, demonstrating its robustness and strong generalization capabilities. Original NWM requires more than 155 s to plan an action trajectory and therefore cannot support responsive real-world replanning.

\begin{table}[H]
    \centering
    \caption{Real-world navigation success rates}
    \label{tab:real-world-results}
    \small
    \resizebox{0.6\linewidth}{!}{%
    \begin{tabular}{lcccc}
        \toprule
        Method & Image-Goal & Language-Goal & Point-Goal\\
        \midrule
        NoMaD   & 40\% & -- & --\\
        OmniVLA & 52\% & 56\% & \textbf{76\%} \\
        NWM-Shortcut & 48\% & 32\% & 68\%\\
        Ours    & \textbf{80\%} & \textbf{68\%} & \textbf{76\%}\\
        \bottomrule
    \end{tabular}%
    }
\end{table}

\section{Limitations}

Our current system is mainly limited to short-range goal-directed navigation. In particular, long-horizon instruction-conditioned navigation remains challenging, as it requires both extended foresight and more complex semantic reasoning than the current world model provides. A promising future direction is to combine the predictive capability of world models with the high-level semantic reasoning ability of VLMs.

\section{Conclusion}

We present a lightweight navigation world model and a unified planning-based framework for multi-modal goal navigation. Our approach addresses a key limitation of prior navigation world models by replacing slow multi-step generation and autoregressive prediction with a one-step generation paradigm built on a 3D U-Net with spatial-temporal attention, enabling efficient future prediction while preserving strong predictive quality. Building on this world model, we further incorporate pretraining and anchor-based planning, which together improve performance and make it possible to support image-goal, language-goal, and point-goal navigation within a single framework. Extensive closed-loop experiments in both simulation and real-world environments demonstrate that the proposed approach achieves strong navigation performance and practical model-based planning across diverse tasks and settings.

\clearpage

\acknowledgments{We thank the anonymous reviewers for their valuable feedback. This work is supported in part by Beijing Natural Science Foundation (Grant No. L252095), in part by National Natural Science Foundation of China (Grant No. 62633013).}

\bibliography{reference}

\clearpage
\appendix
\section*{Appendix A. Qualitative Visualization of Generation Results}

\begin{figure}[H]
    \centering
    \includegraphics[width=0.95\linewidth]{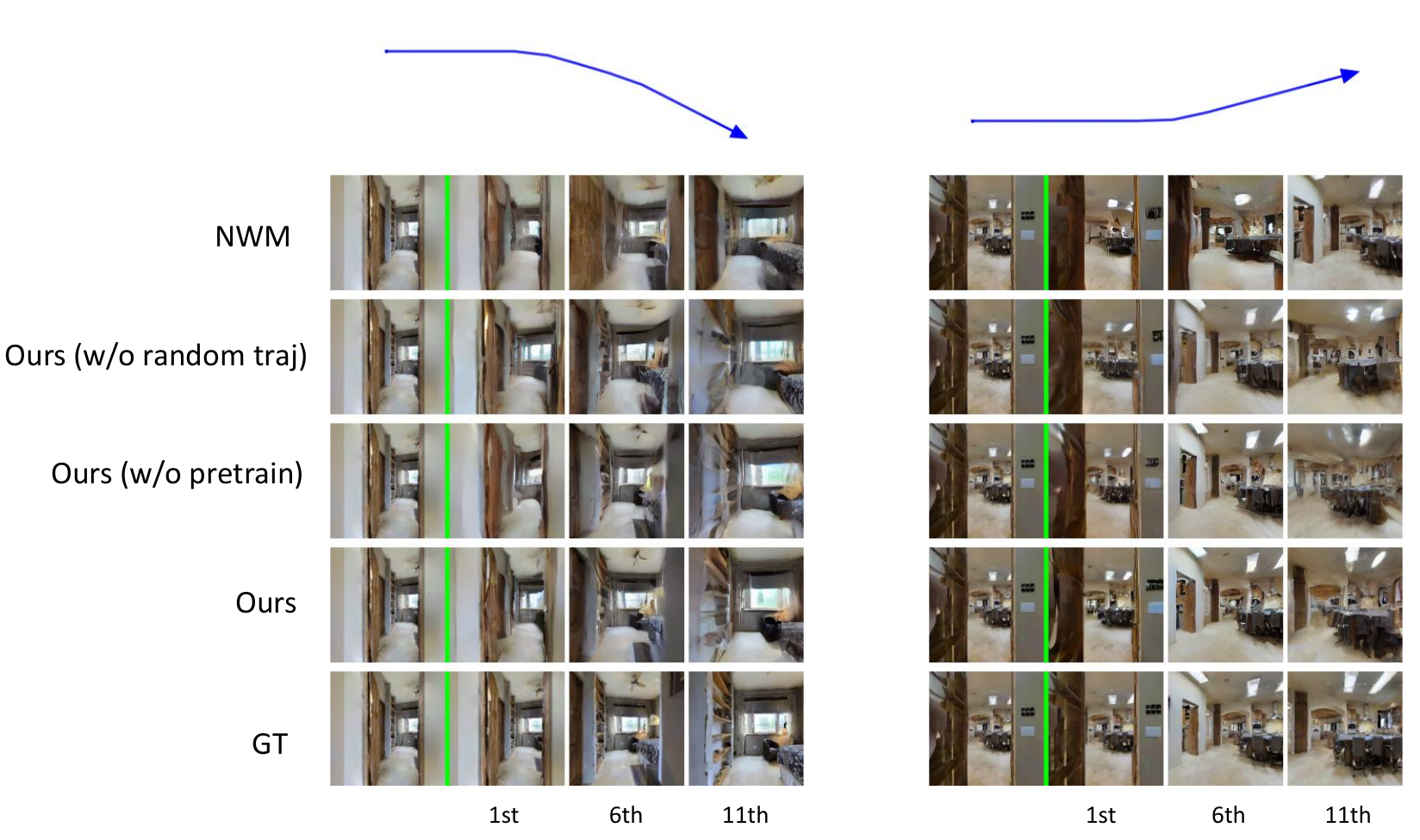}
    \caption{Qualitative comparison of our method against baselines. Our model generates higher-quality frames than NWM and more closely matches the ground truth. The visualization also highlights the benefits of random trajectory sampling and pretraining on public datasets.}
    \label{fig:generation_performance}
\end{figure}

\noindent Figure~\ref{fig:generation_performance} provides a qualitative comparison between our method and the baselines. NWM struggles with precise action execution, and its autoregressive generation process accumulates errors over time, leading to severe position drift from the ground truth. In contrast, the proposed method mitigates these issues by generating all future frames in a single forward pass. The incorporation of temporal attention allows the model to better capture temporal dependencies, resulting in higher-fidelity long-sequence generation. Additionally, random trajectory sampling together with pretraining on large-scale public datasets further improves the visual quality and fine details of the generated images.

\clearpage
\section*{Appendix B. CEM Optimization and Modality-Specific Loss Functions}

We provide a more formal description of the planner used in our navigation system. We combine the Cross-Entropy Method (CEM) with the world model for navigation through the following loop:
\begin{enumerate}
    \item \textbf{Initialize $N$ candidate action trajectories.} At time step $\tau$, let the current observation be $s_{\tau}$. We denote the action predicted at the current step by $a_{\tau}$, where each action is parameterized as a horizon-length sequence of $H$ waypoints. Accordingly, the $i$-th candidate action trajectory is written as
    \begin{equation}
        a_{\tau}^{(i)} = \left[p_{1}^{(i)}, p_{2}^{(i)}, \dots, p_{H}^{(i)}\right].
    \end{equation}
    where each waypoint is represented as $p_{h}^{(i)} = (x_{h}^{(i)}, y_{h}^{(i)}, \phi_{h}^{(i)})$.
    The initial sampling distribution is constructed from the anchor-based initialization described in the main paper, where structured trajectory seeds are generated from fixed combinations of linear and angular velocities to improve action-space coverage under a limited sample budget.

    \item \textbf{Predict future observations with the world model.} Conditioned on $s_{\tau}$ and $a_{\tau}^{(i)}$, the world model predicts a sequence of future observations:
    \begin{equation}
        \hat{s}_{\tau+1:\tau+H}^{(i)} = f_{\theta}(s_{\tau}, a_{\tau}^{(i)}) =
        \left[\hat{s}_{\tau+1}^{(i)}, \hat{s}_{\tau+2}^{(i)}, \dots, \hat{s}_{\tau+H}^{(i)}\right].
    \end{equation}

    \item \textbf{Score each candidate trajectory.} The planner assigns each candidate a scalar loss value
    \begin{equation}
        L(a_{\tau}^{(i)}),
    \end{equation}
    where the specific form of $L(a_{\tau}^{(i)})$ depends on the goal modality and is computed from the predicted future observations. Detailed definitions for image-goal, language-goal, and point-goal navigation are given below.

    \item \textbf{Select the top-$K$ elite trajectories.} Based on the loss values of all candidate trajectories, CEM selects the elite set
    \begin{equation}
        \mathcal{E}_m = \operatorname{TopK}\left(\left\{a_{\tau}^{(i)}\right\}_{i=1}^{N}, K\right),
    \end{equation}
    where $m$ denotes the CEM iteration index, and $\operatorname{TopK}$ returns the $K$ trajectories with the lowest loss values.

    \item \textbf{Update the sampling distribution and re-sample.} The sampling distribution is updated from the elite set as
    \begin{align}
        \mu_{m+1} &= \frac{1}{K} \sum_{a_{\tau}\in\mathcal{E}_m} a_{\tau}, \\
        \Sigma_{m+1} &= \frac{1}{K} \sum_{a_{\tau}\in\mathcal{E}_m} (a_{\tau}-\mu_{m+1})(a_{\tau}-\mu_{m+1})^{\top}.
    \end{align}
    New candidates are then re-sampled from the updated distribution, and the optimization returns to Step 2 to predict future observations for the new candidates and re-evaluate them.
\end{enumerate}
This sampling--prediction--scoring--selection--update loop is repeated until convergence. After $M$ CEM iterations, the planner selects
\begin{equation}
    a_{\tau}^{*} = \arg\min_{a_{\tau}^{(i)}} L(a_{\tau}^{(i)}),
\end{equation}
as the final optimized action for the current control step.

\paragraph{Image-goal loss.}
For image-goal navigation, let the current visual subgoal be $g_{\mathrm{img}}$. We score each candidate trajectory by comparing the predicted future observations against the subgoal image using LPIPS:
\begin{equation}
    L_{\mathrm{img}}(a_{\tau}^{(i)}) =
    \operatorname{LPIPS}\left(\hat{s}_{\tau+H}^{(i)}, g_{\mathrm{img}}\right).
\end{equation}
This objective favors trajectories whose predicted future contains an observation that is perceptually close to the desired visual target.

\paragraph{Language-goal loss.}
For language-goal navigation, let the language instruction be $g_{\mathrm{lang}}$. We use SigLIP to compute the image-text alignment score between each predicted future observation and the language goal, and define the loss as
\begin{equation}
    L_{\mathrm{lang}}(a_{\tau}^{(i)}) =
    - 
    \operatorname{Sim}_{\mathrm{SigLIP}}\left(\hat{s}_{\tau+H}^{(i)}, g_{\mathrm{lang}}\right),
\end{equation}
where $\operatorname{Sim}_{\mathrm{SigLIP}}(\cdot,\cdot)$ denotes the SigLIP similarity score. Minimizing this objective encourages the planner to choose trajectories whose predicted future observations are semantically aligned with the language instruction.

\paragraph{Point-goal loss.}
For point-goal navigation, let $g_{\mathrm{point}}$ be the target coordinate and let $p_H^{(i)}$ denote the terminal point of candidate trajectory $a_{\tau}^{(i)}$. We define the point-goal objective as
\begin{equation}
    L_{\mathrm{point}}(a_{\tau}^{(i)}) =
    \left\| p_H^{(i)} - g_{\mathrm{point}} \right\|_2
    - \lambda_{\mathrm{safe}} \cdot
    \operatorname{Mean}\!\left(
    D\!\left(\hat{s}_{\tau+H}^{(i)}\right)[\mathcal{C}]\right),
\end{equation}
where $D(\cdot)$ denotes the monocular depth estimator, $\mathcal{C}$ denotes the central image region, and $\lambda_{\mathrm{safe}}$ is a weighting coefficient. The first term encourages progress toward the target coordinate, while the second term acts as an explicit safety-oriented regularizer: larger predicted depth in the central region indicates more free space ahead and therefore reduces the loss, discouraging trajectories that move too close to obstacles.

The overall planner uses the modality-specific loss corresponding to the current task:
\begin{equation}
    L(a_{\tau}^{(i)}) \in
    \left\{
    L_{\mathrm{img}}(a_{\tau}^{(i)}),
    L_{\mathrm{lang}}(a_{\tau}^{(i)}),
    L_{\mathrm{point}}(a_{\tau}^{(i)})
    \right\}.
\end{equation}

\end{document}